\documentclass[conference]{IEEEtran}
\IEEEoverridecommandlockouts

\usepackage{cite}
\usepackage{amsmath,amssymb,amsfonts}
\usepackage{algorithmic}
\usepackage{graphicx}
\usepackage{textcomp}
\usepackage{multirow}
\usepackage{xcolor}
\usepackage{url}
\usepackage{bm}
\usepackage[top=.75in, bottom=0.75in, left=0.75in, right=0.75in]{geometry}

\newcommand{\modelName}{BikeVAE-GNN}

\title{BikeVAE-GNN: A Variational Autoencoder-Augmented Hybrid Graph Neural Network for Sparse Bicycle Volume Estimation}
\author{
  \IEEEauthorblockN{Mohit Gupta, Debjit Bhowmick, Ben Beck}
  \IEEEauthorblockA{\textit{School of Public Health and Preventive Medicine, Monash University, Melbourne, Australia}}
  
  \thanks{
  The CYCLED (CitY-wide biCycLing Exposure modelling) Study is funded by ARC DP210102089. Mohit Gupta’s PhD scholarship and Debjit Bhowmick are supported by ARC DP210102089. Ben Beck is supported by ARC FT210100183.
  
    This paper has been accepted for publication in the Proceedings of the $28^{th}$ IEEE International Conference on Intelligent Transportation Systems (ITSC 2025). This is the author's version of the work.
    }
}

\begin{document}

\newgeometry{%
  top=1in, 
  bottom=0.75in, 
  left=0.75in, 
  right=0.75in
}
\maketitle

\begin{abstract}
Accurate link-level bicycle volume estimation is essential for informed urban and transport planning but it is challenged by extremely sparse count data in  urban bicycling networks worldwide.
We propose \textbf{\modelName}, a novel dual-task framework augmenting a Hybrid Graph Neural Network (GNN) with Variational Autoencoder (VAE) to estimate Average Daily Bicycle (ADB) counts, addressing sparse bicycle networks. 
The Hybrid-GNN combines Graph Convolutional Networks (GCN), Graph Attention Networks (GAT), and GraphSAGE to effectively model intricate spatial relationships in sparse networks while VAE generates synthetic nodes and edges to enrich the graph structure and enhance the estimation performance. 
\modelName{} simultaneously performs - regression for bicycling volume estimation and classification for bicycling traffic level categorization. 
We demonstrate the effectiveness of \modelName{} using OpenStreetMap data and publicly-available bicycle count data within the City of Melbourne - where only 141 of 15,933 road segments have labeled counts (resulting in 99\% count data sparsity).
Our experiments show that \modelName{} outperforms machine learning and baseline GNN models, achieving a mean absolute error (MAE) of 30.82 bicycles per day, accuracy of 99\% and F1-score of 0.99. 
Ablation studies further validate the effective role of Hybrid-GNN and VAE components. 
Our research advances bicycling volume estimation in sparse networks using novel and state-of-the-art approaches, providing insights for sustainable bicycling infrastructures. 
\end{abstract}



\section{Introduction}
Bicycling is a sustainable mode of transportation and plays a vital role in reducing traffic congestion, lowering carbon emissions, improving public health and enhancing mobility of diverse populations \cite{wen2008inverse, lindsay2011moving}. 
Accurate link-level bicycling volume estimation, that is estimating the bicycling volumes on individual road-segments across a study area, is critical for informed urban and transport planning, evaluating the effect of bicycling infrastructure, optimizing traffic management and shaping policies to promote bicycling \cite{held2016cycling, bhowmick2023systematic}. 
However, this task is severely challenged by extreme count data sparsity and lack of comprehensive spatial and temporal coverage of count datasets in urban bicycle networks worldwide \cite{winters2017cycling, roy2019correcting}.

Existing methodologies in motorized transportation leverage dense sensor networks and abundant labeled data to achieve high prediction accuracy \cite{jankovic2021traffic, fan2020deep, scats2025}. 
In contrast, the bicycling domain lacks such comprehensive data which necessitates innovative approaches to overcome sparsity and enhance prediction performance. Traditional approaches such as statistical and machine learning models (Direct demand models) struggle in sparse data environments as they rely on sufficient labeled data to generalize and often fail to capture the complex spatial interdependencies in transportation networks \cite{jiang2022graph}. 
Direct demand models \cite{anderson2006direct} are typically trained on data from a small number of sensor locations from high bicycling activity areas and produce biased estimations that poorly generalize to networks with low or uncounted segments.
Recent advances in deep learning, particularly Graph Neural Networks (GNNs) \cite{jiang2022graph} have been effective for motorized traffic prediction by exploiting road network connectivity but their application to bicycling networks is constrained by extreme data sparsity, discussed in section~\ref{subsection: GNN_motor_transport}.

To address this gap, we build upon our prior work \cite{gupta2024evaluating} that systematically evaluated the effects of data sparsity on GNNs for link-level bicycle volume estimation and demonstrated that standard GNN architecture experiences performance degradation under extreme data sparsity. 
In this paper, we introduce \modelName{}, a dual-task Hybrid GNN augmented with Variational Autoencoder (VAE) for accurate estimation of Average Daily Bicycle (ADB) counts across bicycling road network characterized by extreme bike volume/count data sparsity. Our contributions are:
 \begin{itemize}
    \item \textbf{Hybrid-GNN}, a novel framework that integrates GCN, GAT and GraphSAGE to capture complex spatial dependencies in sparse networks.
    \item \textbf{VAE-based data augmentation} generating synthetic nodes and edges to effectively enrich the graph structure for improved model generalization and training.
    \item \textbf{Dual-task learning} for simultaneous regression to estimate continuous bicycling volumes and classification to categorize bicycling traffic levels.
    \item \textbf{Extensive experimentation and ablation studies} to validate \modelName{}'s superior performance against 
    \restoregeometry
    widely-used machine learning models and baseline GNN.
\end{itemize}
\begin{figure*}[t]
    \centering
    \includegraphics[width=1\linewidth]{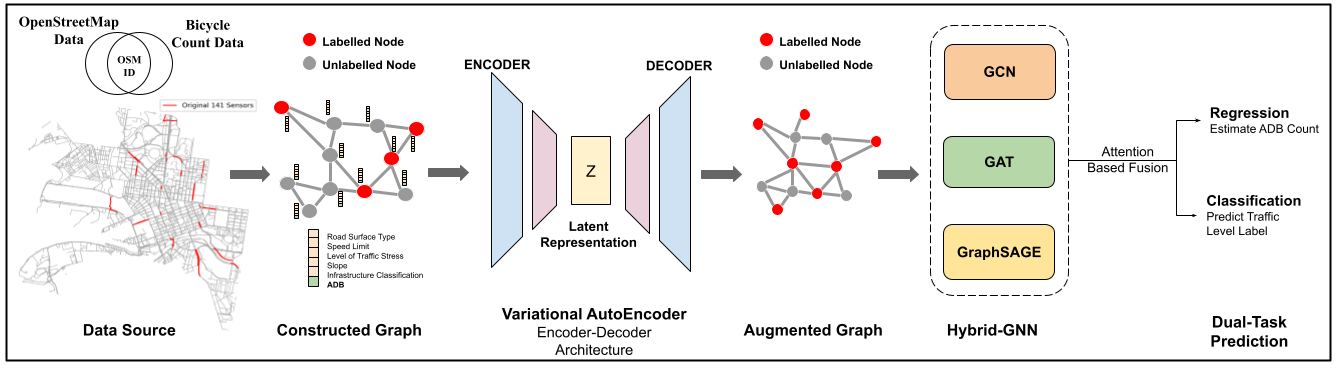}
    \caption{Architecture of the \modelName{} framework. 
    The workflow begins with constructing a node-centric graph by combining OpenStreetMap infrastructure data and City of Melbourne bicycle count data (only 141 of  15,933 segments labeled). A VAE augments this graph by generating synthetic nodes and edges via an encoder-decoder process to produce a denser augmented graph. The Hybrid-GNN module processes the augmented graph through parallel GCN, GAT, and GraphSAGE branches.  An attention-based fusion layer aggregates their embeddings and pass to dual task-specific heads - regression of ADB counts and classification of bicycling traffic levels.}
    \label{fig:architecture}
\end{figure*}

\section{Related Work}
\subsection{Bicycle Volume Estimation Methods}
Early bicycle volume estimation studies adopted generalized linear models such as Poisson, negative‐binomial or zero‐inflated variants, to estimate bicycle counts \cite{griswold2011pilot, miranda2011weather}. Although these models provide interpretable results, their linearity assumptions limit their ability to capture complex traffic patterns.
More recent studies have used ML models such as Support Vector Regression (SVR) and Random Forest (RF) \cite{miah2023estimation, tao2024use} to model non-linear relationships in bicycle traffic patterns. However, these methods require abundant labeled training data to generalize and therefore, struggled with sparse data. Also, they typically assume that road segments as independent and identically distributed observations and ignore the critical inherent spatial dependencies in the topological structure of a road network \cite{nikparvar2021machine}. 

\subsection{Challenges of Data Sparsity in Bicycling Networks}
Data sparsity remains a pervasive challenge in urban bicycle networks worldwide and limits the effectiveness of volume estimation models. 
Unlike motorized transportation networks which benefit from extensive sensor coverage using loop detectors, automated traffic counters and cameras \cite{gu2017traffic}, bicycling networks have historically been overlooked and there is a significant gap due to limited sensor deployment and inconsistent data collection methodologies which impede the ability to accurately estimate and analyze bicycling patterns.
Studies from various cities such as Melbourne \cite{gupta2024evaluating}, Krakow \cite{pogodzinska2020bicycle}, and Berlin \cite{kaiser2025counting} highlight the issue of sparse bicycling data. 
In Melbourne, Australia, less than $1\%$ of 15,933 road segments have the bicycle count data and Berlin has only 35 counting points for more than 1,300 km of cycling network\cite{kaiser2025counting}, reflecting a common issue in cities with sparse sensor networks \cite{gupta2024evaluating}. 
A global analysis of 62 cities further revealed that even well-developed bicycle networks struggle with sparse data which affects infrastructure growth and planning \cite{szell2022growing}
These examples underscore the importance of models designed to perform robustly under sparse data conditions, a significant gap in existing bicycling research.

\subsection{Graph Neural Networks in Transportation} \label{subsection: GNN_motor_transport}
GNNs have emerged as powerful methods to effectively model and capture spatial dependencies inherent within transportation networks 
GNNs have shown promising results in motorized traffic prediction by explicitly modeling road network topology and outperform traditional statistical and machine learning models in motorised transportation domain \cite{jiang2022graph}. 
Foundational architectures such as GCN \cite{kipf2016semi}, GAT \cite{velivckovic2017graph} and GraphSAGE \cite{hamilton2017inductive} have demonstrated significant performance in capturing node-level relationships in transportation networks by aggregating feature information from neighborhood nodes.
They have been successfully adapted for transportation applications such as predicting traffic congestion levels, travel demand forecasting, incident detection tasks \cite{zhang2020spatial, ye2020build}. 
Building upon these foundational architecture, GNN models have been developed with spatio-temporal extensions such as Diffusion Convolutional Recurrent Neural Network (DCRNN) \cite{li2017diffusion}, Spatio-Temporal Graph Convolutional Networks (ST-GCN) \cite{yu2017spatio} further showing remarkable performance in traffic prediction tasks over large-scale urban networks.

\subsection{Graph Generative Models for Sparsity Mitigation}
The challenge of data sparsity in modeling networks with GNNs has prompted recent advances in graph representation learning to explore generative approaches to artificially augment and enrich datasets. 
Among these models, especially graph generative models deploying VAEs have shown promise in synthesizing structurally coherent graphs while preserving essential network characteristics.
VAE for graphs such as GraphVAE \cite{simonovsky2018graphvae} have demonstrated their ability to generate small-scale graphs by learning latent node and edge distributions, thereby offering an effective tool for augmenting sparse datasets.
More recent advances including SIG-VAE \cite{hasanzadeh2019semi} and Dirichlet-GVAE \cite{li2020dirichlet} have extended to imporve the quality and diversity of synthesized graph data by incorporating richer probabilistic formulations.
Another notable class of graph generative models includes adversarial approaches such as NetGAN \cite{bojchevski2018netgan} which uses adversarial training frameworks to produce realistic network topologies through the emulation of random walks to capture  accurate preservation of higher-order network patterns. 
While these generative methods have achieved notable success in diverse domains such as molecular structure generation and social network analysis, they have not been used in the transportation domain especially in the context of bicycle networks. These methods prioritize global topology preservation over domain-specific semantics and functional attributes such as connectivity, infrastructure type and traffic stress levels which directly impact the modeling performance \cite{jin2020graph}. 

\textbf{\modelName{}} framework proposed in this study directly addresses these limitations by integrating a hybrid GNN architecture with a tailored VAE-based augmentation approach. 
Specifically, our Hybrid-GNN combines GCN for local neighborhood feature aggregation, GAT for adaptive neighbor weighting and GraphSAGE for inductive node embedding learning to effectively capture the multi-scale spatial patterns within extremely sparse bicycle networks. 
Additionally, our VAE component specifically generates synthetic nodes with realistic and domain-appropriate bicycling volume attributes based on OSM infrastructure features to enhance data density and robustness without sacrificing the semantic integrity of the network.
To the best of our knowledge, this is the first framework to integrate hybrid-GNNs with domain-tuned VAE-based semantic data augmentation to perform link-level bicycle volume modeling in extreme data sparsity, addressing a critical gap identified in our prior analysis of GNN limitations under sparsity \cite{gupta2024evaluating}.

\section{Methodology}

\subsection{Problem Formulation}

We formulate bicycling volume estimation as a graph-based dual-task learning problem on a road network represented as an undirected graph $\mathcal{G} = (\mathcal{V}, \mathcal{E}, \mathbf{X})$, where $\mathcal{V}$ denotes the set of $N$ road segments (nodes), $\mathcal{E} \subseteq \mathcal{V} \times \mathcal{V}$ represents intersections (edges) and $\mathbf{X} \in \mathbb{R}^{N \times d}$ contains $d$-dimensional feature vectors for each node.
Given extreme count data sparsity with only $\mathbf{K} << \mathcal{N}$ labeled nodes, we simultaneously learn: (1) a regression task $f_{\text{reg}}$ estimating continuous ADB counts $\hat{y}_i^{\text{reg}} \in \mathbb{R}^+$ using neighborhood-aware features, and (2) a classification task $f_{\text{clf}}$ predicting bicycling traffic levels $\hat{y}_i^{\text{clf}} \in \{1,...,5\}$ (Very Low to Very High).
The joint optimization objective combines both tasks:
\begin{equation}
    \mathcal{L} = \alpha \cdot \underbrace{\frac{1}{K}\sum_{i=1}^K (y_i^{\text{reg}} - \hat{y}_i^{\text{reg}})^2}_{\text{MSE}} + (1-\alpha) \cdot \underbrace{\frac{1}{K}\sum_{i=1}^K \text{CE}(y_i^{\text{clf}}, \hat{y}_i^{\text{clf}})}_{\text{Cross-Entropy}}
    \label{eq:loss}
\end{equation}
where $\alpha$ balances task importance.
To address high data sparsity ($\frac{K}{N}$), we augment $\mathcal{G}$ with synthetic nodes $\tilde{\mathbf{X}}$ generated via a VAE and edges $\tilde{\mathcal{E}}$ created through cosine similarity thresholding, forming an enriched graph $\mathcal{G}'$ for robust training of Hybrid-GNN architecture.


\subsection{Hybrid-GNN Architecture}  \label{subsection: hybrid-gnn}
Our Hybrid-GNN architecture integrates three GNN variants (as shown in Figure~\ref{fig:architecture}) to model spatial dependencies for link-level bicycle volume estimation in bicycling network with sparse count data:

\textbf{\textit{1. Graph Convolutional Network (GCN):}}
Captures local neighborhood patterns through spectral graph convolutions.
\begin{equation}
    \mathbf{h}_{i}^{(l, \text{GCN})} = \sigma \left( \sum_{j \in \mathcal{N}(i) \cup \{i\}} \frac{1}{\sqrt{d_i d_j}} \mathbf{W}_{\text{GCN}}^{(l)} \mathbf{h}_{j}^{(l-1)} \right)
\end{equation}
where, $\mathbf{h}_{i}^{(l, \text{GCN})}$ is the hidden representation of node $v_i$ (the $i$-th road segment) at layer $l$, $\mathcal{N}(i) \cup \{i\}$ includes $v_i$ and its neighbors (connected via edges $\mathcal{E}$), $d_i = |\mathcal{N}(i)| + 1$ is the degree accounting for the self-loop, $\mathbf{W}_{\text{GCN}}^{(l)}$ is a learnable weight matrix, and $\sigma$ is ReLU activation function. The normalization factor $\frac{1}{\sqrt{d_i d_j}}$ ensures stable gradient flow by balancing contributions from nodes with varying degrees, a key advantage in handling the heterogeneous connectivity of urban bicycle networks. GCN's spectral approach transforms graph signals into a lower-dimensional space, making it suitable for extracting local patterns from the VAE-augmented graph.

\textbf{\textit{2. Graph Attention Network (GAT):}}  
Models attention-based dependencies and assigns adaptive weights to neighbors based on their relevance which is crucial for capturing heterogeneous traffic patterns in sparse bicycle networks. Unlike GCN’s uniform aggregation, GAT prioritizes influential neighbors which makes it effective for urban networks with varying connectivity. 
\begin{equation}
    \alpha_{ij}^{(l)} = \phi_j \left( \text{LeakyReLU} \left( \mathbf{a}^T [\mathbf{W}_{\text{GAT}}^{(l)} \mathbf{h}_{i}^{(l-1)} \| \mathbf{W}_{\text{GAT}}^{(l)} \mathbf{h}_{j}^{(l-1)}] \right) \right)
\end{equation}
\begin{equation}
    \mathbf{h}_{i}^{(l,\text{GAT})} = \sigma \left( \sum_{j \in \mathcal{N}(i) \cup \{i\}} \alpha_{ij}^{(l)} \mathbf{W}_{\text{GAT}}^{(l)} \mathbf{h}_{j}^{(l-1)} \right)
\end{equation}
where, $\mathbf{h}_{i}^{(l,\text{GAT})}$ is the hidden representation of node $v_i$ at layer $l$, $\mathcal{N}(i) \cup \{i\}$ includes $v_i$ and its neighbors, $\phi_j$ is the softmax function, $\mathbf{W}_{\text{GAT}}^{(l)}$ is a learnable weight matrix, $\alpha_{ij}^{(l)}$ is the attention coefficient between nodes $i$ and $j$. $\alpha_{ij}^{(l)}$ is the attention coefficient, $\mathbf{a}$ is a learnable attention vector, $\|$ shows concatenation, $\text{LeakyReLU}$ is used to stabilize training. 
This mechanism allows GAT to weigh contributions from high-traffic neighbors more heavily, enhancing the model’s ability to infer volumes in unlabeled regions of the VAE-augmented graph.

\textbf{\textit{3. GraphSAGE:}}
Enables inductive learning through neighborhood sampling and aggregation and improves scalability and generalization to unseen nodes in the augmented graph.  

\begin{equation}
\mathbf{h}_i^{(l,\text{SAGE})} = \sigma\Bigl(
    \mathbf{W}_{\text{SAGE}}^{(l)} 
    \cdot 
    \text{CONCAT}\bigl(
        \underset{j \in \mathcal{S}(i)}{\text{MEAN}}(\mathbf{h}_j^{(l-1)})
        \mathbf{h}_i^{(l-1)}, 
    \bigr)
\Bigr)
\end{equation}
where, $\mathbf{h}_{i}^{(l,\text{GraphSAGE})}$ is the hidden representation of node $v_{i}$ at layer $l$, $\mathbf{W}_{\text{GraphSAGE}}^{(l)}$ are learnable weights and $\mathcal{N}(i)$ represents the neighborhood of node $v_i$, $\mathcal{S}(i)$ is a randomly sampled neighbor subset and \( \sigma \) is ReLU. 
GraphSAGE samples a fixed number of neighbors (e.g., 10–20) to reduce computational complexity which makes it efficient for large-scale urban networks. This inductive approach allows the model to generalize to newly added nodes or edges in the VAE-augmented graph. 

\textbf{\textit{Architectural Integration:} }
We proposed two architectural configurations - Parallel and Sequential, to capture multi-scale spatial patterns in sparse bicycle networks. 
In the \textit{Parallel} configuration, three GNN branches simultaneously process node features \( \mathbf{X} \in \mathbb{R}^{N \times d} \). Each branch consists of a single layer: GCN (\( \mathbf{h}_i^{\text{GCN}}\)), GAT with one attention head (\( \mathbf{h}_i^{\text{GAT}}\)), and GraphSAGE (\( \mathbf{h}_i^{\text{SAGE}} \)). 
This design enables simultaneous learning of localized patterns (GCN), adaptive relationships (GAT), and generalized neighborhood structures (GraphSAGE).
We concatenate outputs from all three components using attention-based fusion to form hybrid node representations:  
\begin{equation}
\mathbf{h}_i^{(l)} = \text{CONCAT}\left(\mathbf{h}_i^{(l,\text{GCN})}, \mathbf{h}_i^{(l,\text{GAT})}, \mathbf{h}_i^{(l,\text{SAGE})}\right)
\end{equation}  
In the \textit{Sequential} configuration, the data are processed through stacked GNN layers to form hybrid node representation:
\begin{equation}
\mathbf{h}_i^{(l)} = \text{GraphSAGE}\bigl(\text{GAT}\bigl(\text{GCN}(\mathbf{h}_i^{(l-1)})\bigr)\bigr)
\end{equation}
Final predictions are computed through task-specific heads:  
\begin{equation}
\hat{y}_i^{\text{reg}} = \mathbf{W}_{\text{reg}} \mathbf{h}_i^{(L)}, \quad \hat{y}_i^{\text{clf}} = \text{softmax}(\mathbf{W}_{\text{clf}} \mathbf{h}_i^{(L)})
\end{equation}  
This design enables simultaneous learning of localized patterns (GCN), adaptive relationships (GAT), and generalized neighborhood structures (GraphSAGE), crucial for handling extremely sparse networks.


\subsection{VAE-Based Augmentation} \label{subsection: vae}

To address extreme sparsity (\( \frac{K}{N} \)) in the bicycle network graph \( \mathcal{G} \), \modelName{} employs a graph-aware VAE framework that generates synthetic nodes and edges to form an augmented graph \( \mathcal{G}' \) while reserving the topological properties of the original bicycling network. The VAE learns the distribution of node features \( \mathbf{X} \in \mathbb{R}^{N \times d} \) to enrich the training data for the hybrid GNN.

The VAE comprises: an encoder mapping input features \( \mathbf{x}_i \) to a 32-dimensional latent space (\( \mathbf{z}_i \sim \mathcal{N}(\mu_i, \sigma_i) \)) via two linear layers (128, 64 units) with ReLU, 
\begin{equation}
        \bm{\mu}_i, \bm{\sigma}_i = \text{MLP}_{\text{enc}}(\mathbf{x}_i), \quad \mathbf{z}_i \sim \mathcal{N}(\bm{\mu}_i, \text{diag}(\bm{\sigma}_i^2))
\end{equation}
and a decoder reconstructing \( \tilde{\mathbf{x}}_i \) through symmetric layers (64, 128 units) with sigmoid output for normalized features. 
\begin{equation}
        \tilde{\mathbf{x}}_i = \text{MLP}_{\text{dec}}(\mathbf{z}_i), \quad p_{ij} = \sigma(\mathbf{z}_i^T \mathbf{W}_{\text{edge}} \mathbf{z}_j)
\end{equation}
Edge probabilities are predicted as \( p_{ij} = \sigma(\mathbf{z}_i^T \mathbf{W}_{\text{edge}} \mathbf{z}_j) \).

We sample synthetic node features from the latent space \( \tilde{\mathbf{z}}_j \sim \mathcal{N}(\bm{0}, \mathbf{I}) \), computing \( \tilde{\mathbf{x}}_j = \text{MLP}_{\text{dec}}(\tilde{\mathbf{z}}_j) \), clipped to [0,1]. Edges \( \tilde{\mathcal{E}} \) connect each synthetic node to the top-5 original labeled nodes where cosine similarity exceeds \( \tau=0.7 \), ensuring connectivity to labeled nodes. 
\begin{equation}
\tilde{\mathcal{E}} = \{(j,i) \mid \text{cos}(\tilde{\mathbf{x}}_j, \mathbf{x}_i) > \tau, i \in \mathcal{V}_{\text{labeled}}\}.
\end{equation}
Pseudo-labels are assigned using the pre-trained hybrid GNN: \( \tilde{y}_j^{\text{reg}} = f_{\text{reg}}(\tilde{\mathbf{x}}_j) \), \( \tilde{y}_j^{\text{clf}} = f_{\text{clf}}(\tilde{\mathbf{x}}_j) \). The VAE is trained for 500 epochs and optimizes a composite loss:
\begin{align}
\mathcal{L}_{\text{VAE}} &= \underbrace{\|\mathbf{x}_i - \tilde{\mathbf{x}}_i\|^2}_{\text{Reconstruction}} \notag + \beta \cdot \underbrace{D_{\text{KL}}(\mathcal{N}(\bm{\mu}_i, \bm{\sigma}_i^2) \| \mathcal{N}(\bm{0}, \mathbf{I}))}_{\text{Regularization}} \notag \\
&\quad + \gamma \cdot \underbrace{\text{BCE}(\mathbf{A}, \mathbf{P})}_{\text{Edge Prediction}}
\end{align}
where, $D_{\text{KL}}$ is Kullback-Leibler divergence, $\text{BCE}$ is Binary Cross-Entropy loss, $\text{A}$ is ground truth and $\text{P}$ is predicted adjacency matrix respectively. 
While maintaining the structural properties of the original bicycling network, this augmentation strategy increases the effective training set size and enhances hybrid GNN performance enabling robust modeling of sparse bicycle network.

\subsection{Dataset and Preprocessing - Melbourne Case Study}
Our study focuses on the City of Melbourne, a dense urban area with a complex network of roads, bike lanes, and shared paths, covering 37.7 square kilometers and comprising 15,933 individual links. 
Two primary data sources are used: 
\begin{enumerate}
\item \textbf{OpenStreetMap (OSM) Infrastructure Data}: OSM provides a comprehensive open-source data repository of geographical and infrastructural information. 
For this study, OSM data serves as a foundational data and we extract the bicycle network (all roads/paths where bicyclists are permitted to ride) for the
City of Melbourne using the OSMnx library \cite{boeing2017osmnx}. 
OSM provides road network attributes (road surface type, speed limit, slope, infrastructure type, level of traffic stress) via OSMnx. 
Further, we use a bicycle infrastructure classification system \cite{Sustainable_Mobility_and_Safety_Research_Group_Bicycling_infrastructure_classification_2023} and bicycle level of traffic stress (LTS) model \cite{bike_lts} bespoke to Greater Melbourne to appropriately classify the types of bicycle infrastructure and bicycle LTS for each street segment in the network, respectively.
We also supplement missing or inaccurately recorded attributes in OSM, such as slope and speed limits respectively, by spatially merging the OSM network with additional datasets - slope data from Department of Energy, Environment and Climate Action Victoria, 2021 \cite{slope_data} (slope data) and speed limit data from Department of Transport and Planning (DTP) Victoria, 2014 \cite{DTP_speed_data}, ensuring a more comprehensive representation of key features relevant to bicycling volumes on road segments.
\item \textbf{City of Melbourne (CoM) Bicycling Count Data}: The count data provided by DTP Victoria \cite{dtp_count_data}, includes daily bicycle counts on roads and streets across City of Melbourne, Australia.  This dataset represents the observed bicycling activity and serves as the ground-truth data to derive ADB counts~\eqref{eq:ADB_eq}. 
However despite its value, it exhibits significant sparsity, with only 141 segments (out of 15,933) having recorded bicycle counts - approximately 99\% data sparsity.

\end{enumerate}

For each segment with available counts, we compute the ADB count by aggregating and averaging daily counts over the available period. The observed ADB counts range from 2 to 818 average bicycles per day across the 141 segments. These ADB values serve as regression targets. 
\begin{equation}\label{eq:ADB_eq}
    ADB_i=\left\lceil\frac{\sum_{j=1}^{n_i}{bicycle\_trip\_count}_{i j}}{n_{i}}\right\rceil
\end{equation}

To support the classification task, ADB counts are discretized into five quantile-based bicycling traffic level categories: Very Low, Low, Medium, High, and Very High Traffic. 

This stratification ensures balanced representation across classes. 
All features are normalized prior to modeling. 
The final dataset integrates OSM-based features with CoM ADB and CoM traffic labels for each of the 141 labeled road segments, providing both regression and classification targets for subsequent modeling.
\subsection{Graph Construction}
\label{graph_construction}
The City of Mebourne's bicycle network is modeled as an undirected graph, $\mathcal{G}$. Unlike OpenStreetMap’s edge-centric convention, we represent road segments as nodes and intersections as edges to enable node-level bicycling volume estimation as shown in Figure ~\ref{fig:architecture}. 
The feature matrix incorporates preprocessed attributes including road type, slope, speed limit, level of traffic stress and infrastructure type. 
 \( \mathcal{G} \) serves as input to the VAE for augmentation, producing \( \mathcal{G}' \) (\ref{subsection: vae}), and supports hybrid GNN training (\ref{subsection: hybrid-gnn}) for sparse volume prediction.

\subsection{Training Procedure}
The training procedure for all models is designed to ensure robust evaluation and fair comparison across baselines, GNN variants, and the proposed Hybrid-GNN with VAE augmentation. 
For both regression and classification tasks, we employ a 5-fold cross-validation strategy, randomly partitioning the 141 labeled nodes into training and test sets with a 70/30 split in each fold. 
Stratified splits are used for classification to preserve the distribution of bicycling traffic level classes.

For traditional machine learning models (RF, SVR, MLP), model training and hyperparameter tuning are performed using scikit-learn, with standard settings unless otherwise specified. 
For all GNN-based models including GCN, GAT, GraphSAGE and both parallel and sequential Hybrid-GNN architectures, training is conducted in PyTorch Geometric \cite{fey2019fast}. 
Each model is trained for up to 500 epochs using the Adam optimizer with a learning rate of 0.01, a batch size of 64, and dropout rate of 0.5 to mitigate overfitting. 
Early stopping is applied if validation loss does not improve for 50 consecutive epochs.

The VAE is trained separately on node features for 100 epochs with a latent dimension of 32, using the Adam optimizer and a learning rate of $1 \times 10^{-3}$. 
Synthetic nodes generated by the VAE are assigned pseudo-labels using pre-trained Hybrid-GNN models and are integrated into the training set for subsequent model runs.

During each fold, model performance is evaluated on the held-out test set using mean absolute error (MAE), root mean squared error (RMSE), and $R^2$ for regression, and accuracy, precision, recall, and F1-score for classification. 
For regression, all error metrics are reported in units of bicycles per day, corresponding to the ADB counts.
All experiments are conducted with fixed random seeds for reproducibility and results are averaged across folds. 

\section{Results and Discussion}


\begin{table*}[t]
\centering
\caption{Performance Comparison of Baseline Models and Proposed BikeVAE-GNN}
\label{table:baseline}
\renewcommand{\arraystretch}{1.2} 
\begin{tabular}{l|cccc|cccc}
\hline
\multicolumn{1}{c|}{\multirow{2}{*}{\textbf{Model}}} & \multicolumn{4}{c|}{\textbf{REGRESSION}} & \multicolumn{4}{c}{\textbf{CLASSIFICATION}} \\ \cline{2-9} 
\multicolumn{1}{c|}{} & \textbf{MAE} & \textbf{RMSE} & \textbf{MAPE (\%)} & \textbf{R\textsuperscript{2}} & \textbf{Accuracy (\%)} & \textbf{Precision (\%)} & \textbf{Recall (\%)} & \textbf{F1-Score} \\ \hline
SVM & 158.12 & 202.62 & 8.18 & 0.01 & 0.93 & 0.95 & 0.93 & 0.93 \\
RF & 92.41 & 134.43 & 6.65 & 0.25 & 0.94 & 0.96 & 0.93 & 0.93 \\
MLP & 208.86 & 252.10 & 6.63 & 0.57 & 0.91 & 0.90 & 0.86 & 0.89 \\ \hline
Baseline GCN & 150.56 & 189.05 & 5.44 & 0.63 & 0.95 & 0.96 & 0.95 & 0.95 \\
Baseline GAT & 137.91 & 176.53 & 5.17 & 0.62 & 0.93 & 0.95 & 0.93 & 0.93 \\
Baseline GraphSAGE & 105.10 & 137.74 & 5.59 & 0.73 & 0.93 & 0.94 & 0.93 & 0.93 \\ \hline
HybridParallelGNN & 37.91 & 61.50 & 2.39 & 0.92 & 0.97 & 0.97 & 0.97 & 0.96 \\
HybridSequentialGNN & 65.20 & 85.22 & 2.83 & 0.86 & 0.96 & 0.97 & 0.96 & 0.96 \\ \hline
\textbf{HybridParallelGNN + VAE} & \textbf{30.82} & \textbf{50.18} & \textbf{0.16} & \textbf{0.91} & \textbf{0.99} & \textbf{0.99} & \textbf{0.99} & \textbf{0.99} \\
HybridSequentialGNN + VAE & 35.28 & 53.28 & 0.17 & 0.90 & 0.97 & 0.98 & 0.97 & 0.97
\end{tabular}
\end{table*}

\begin{table*}[t]
\centering
\caption{Ablation Study for \modelName{}}
\label{table:ablation_study}
\renewcommand{\arraystretch}{1.2} 
\centering
\begin{tabular}{l|cccc|cccc}
\hline
\multicolumn{1}{c|}{\textbf{Model}} & \textbf{MAE} & \textbf{RMSE} & \textbf{MAPE (\%)} & \textbf{R\textsuperscript{2}} & \textbf{Accuracy (\%)} & \textbf{Precision (\%)} & \textbf{Recall (\%)} & \textbf{F1-Score} \\ \hline
BikeVAE-GNN: Without GAT & 35.50 & 55.60 & 0.20 & 0.89 & 0.98 & 0.98 & 0.98 & 0.98 \\
BikeVAE-GNN: Without GraphSAGE & 33.20 & 52.30 & 0.18 & 0.90 & 0.98 & 0.99 & 0.98 & 0.98 \\
BikeVAE-GNN: Without VAE & 45.10 & 75.40 & 0.30 & 0.85 & 0.96 & 0.97 & 0.96 & 0.96
\end{tabular}
\end{table*}

\subsection{Baseline Results}
Baseline models faced significant challenges in modeling Melbourne’s bicycle network due to extreme data sparsity (\( \frac{K}{N} \approx 0.009 \)), with only 141 labeled segments out of 15,933. Machine learning (ML) baselines—Support Vector Machine (SVM, MAE=158.12, RMSE=202.62), Random Forest (RF, MAE=92.41, RMSE=134.43), and Multi-Layer Perceptron (MLP, MAE=208.86, RMSE=252.10) rely solely on node features, failing to capture spatial dependencies inherent in the network (Table \ref{table:baseline}). 
Their classification performance was moderate (Accuracy=0.91–0.94, F1-Score=0.89–0.93), reflecting challenges in distinguishing bicycling traffic level quintiles without structural context. 
GNN baselines — Graph Convolutional Network (GCN, MAE=150.56), Graph Attention Network (GAT, MAE=137.91), and GraphSAGE (MAE=105.10) leveraged the graph \( \mathcal{G} \), improving regression (R²=0.62–0.73) and classification (Accuracy=0.93–0.95) over ML models. 
However, their performance was constrained by the limited labeled data, with GraphSAGE achieving the best baseline regression (MAE=105.10) and GCN the highest classification accuracy (0.95). These results underscore the need for advanced augmentation to address sparsity.

\subsection{VAE Augmentation Hybrid GNN Results}
The \modelName{} framework, specifically HybridParallelGNN + VAE, significantly outperformed all baselines, achieving MAE=30.82, RMSE=50.18, MAPE=0.16\%, and R²=0.91 for regression, and Accuracy=0.99, F1-Score=0.99 for classification (Table \ref{table:baseline}). 
Compared to the best ML baseline (RF, MAE=92.41), \modelName{} reduces MAE by 66.7\%, and against the best GNN baseline (GraphSAGE, MAE=105.10), it reduces MAE by 70.7\%. 
The VAE augments the graph, \( \mathcal{G}' \), generating synthetic features that enhance data robustness, while the hybrid parallel GNN architecture effectively captures spatial dependencies through combined GAT and GraphSAGE aggregations. 
In contrast, HybridSequentialGNN + VAE (MAE=35.28, Accuracy=0.97) performs slightly worse, as sequential processing limits feature integration. 
HybridParallelGNN without VAE (MAE=37.91) shows a 23.0\% higher MAE, highlighting VAE’s role in addressing sparsity. 
The classification metrics demonstrate \modelName{}’s ability to accurately predict bicycling traffic levels, making it a robust solution for urban mobility modeling in data-scarce environments.

\subsection{Ablation Study}
We ablate key components of \modelName{} to quantify their contributions (Table \ref{table:ablation_study}). 
Removing the VAE augmentation degrades performance significantly, increasing MAE from 30.82 to 45.10 and reducing accuracy from 0.99 to 0.96, as the model relies on the sparse \( \mathcal{G} \) (141 nodes) without synthetic features to bridge data gaps. 
This 46.3\% MAE increase underscores VAE’s critical role in generating robust node representations for sparse networks. Omitting GAT raises MAE to 35.50 (15.2\% increase), while removing GraphSAGE results in MAE=33.20 (7.7\% increase), indicating their complementary contributions to feature aggregation. 
GAT’s attention mechanism enhances neighbor weighting, and GraphSAGE’s sampling improves scalability, but their absence is less detrimental than VAE’s as the hybrid architecture retains partial functionality. Classification metrics remain high (Accuracy=0.98) in GAT and GraphSAGE ablations, but drop to 0.96 without VAE, reinforcing VAE’s dominance in enabling accurate bicycling traffic level predictions. 
These results confirm that VAE augmentation and the hybrid GNN architecture are both essential for \modelName{}’s superior performance in sparse urban bicycle networks.

\section{Conclusion}
This study presents \modelName{}, a novel framework integrating VAE augmentation with a hybrid parallel GNN to model bicycle volumes in networks with sparse count data.
Addressing extreme data sparsity (\( \frac{K}{N} \approx 0.009 \)), \modelName{} achieves impressive performance, with MAE=30.82 and Accuracy=0.99, reducing MAE by 66.7\% compared to RF (MAE=92.41) and 70.7\% compared to GraphSAGE (MAE=105.10), as shown in Table \ref{table:baseline}. 
The VAE augments the graphs by generating robust features while the hybrid GNN combines GNN,  GAT and GraphSAGE to capture spatial dependencies effectively. Ablation studies (Table \ref{table:ablation_study}) confirm VAE’s critical role, with its removal increasing MAE by 46.3\% to 45.10, underscoring its importance in sparse settings.

\modelName{} offers significant implications for urban mobility planning, enabling accurate prediction of bicycle volumes and bicycling traffic levels to inform infrastructure optimization and cyclist safety policies. 
Future work includes validating \modelName{} on larger, multi-city datasets, incorporating temporal dynamics to capture seasonal variations, and extending the framework to other transport modes, such as pedestrian or vehicular traffic. 
These advancements will further enhance data-driven urban transportation systems, supporting sustainable and safe mobility.


\bibliographystyle{ieeetr}
\bibliography{references}


\end{document}